\documentclass[a4paper,12pt]{article}
\usepackage{amsmath}
\usepackage{times}
\usepackage{graphicx}
\usepackage{color}
\usepackage{multirow}
\usepackage[authoryear]{natbib}
\usepackage{rotating}
\usepackage{bbm}
\usepackage{latexsym}
\usepackage{url}            

\newcommand{\sgap}{\vspace*{3mm}}
\newcommand{\rlam}{\sqrt{\lambda}}

\title{
\makebox[0pt][l]{\raisebox{1in}[0pt][0pt]{\hspace{-1in}
\parbox{6in}{\small Final m/s version of paper published in
  \emph{Neural Computation} 34(10) 2037-2046 (2022), with an addendum
(arXiv only) added in March 2023.}}}
On Suspicious Coincidences  and \\ Pointwise Mutual Information}

\author{
  Christopher K.~I.~Williams \\
  School of Informatics \\
  University of Edinburgh, UK \\
  \texttt{c.k.i.williams@ed.ac.uk} \\
  }

\date{\today}

\begin{document}

\maketitle

\begin{abstract}
\citet{barlow-85} hypothesized that the co-occurrence of two events
$A$ and $B$ is ‘suspicious’ if $P(A,B) \gg P(A) P(B)$. We first review
classical measures of association for $2 \times 2$ contingency tables,
including Yule's $Y$ \citep{yule-1912}, which depends only on the odds
ratio $\lambda$, and is independent of the marginal probabilities of
the table. We then discuss the mutual information (MI) and pointwise
mutual information (PMI), which depend on the ratio $P(A,B)/P(A)P(B)$,
as measures of association.  We show that, once the effect of the
marginals is removed, MI and PMI behave similarly to $Y$ as functions
of $\lambda$. The pointwise mutual information is used extensively in
some research communities for flagging suspicious coincidences. We
discuss the pros and cons of using it in this way, bearing in mind the
sensitivity of the PMI to the marginals, with increased scores for
sparser events.
\end{abstract}

\citet{barlow-85} hypothesized that ``the cortex behaves like a gifted
detective, noting suspicious coincidences in its afferent input, and
thereby gaining knowledge of the non-random, causally related,
features in its environment''.
More specifically, Barlow wrote (p.\ 40):
\begin{quote}
  The coincident occurrence of two events $A$ and $B$ is `suspicious'
  if they occur jointly more than would be expected from the
  probabilities   of their individual occurrence, i.e.\ the
  coincidence $A \& B$ is suspicious if $P(A \& B) \gg P(A) \times 
  P(B)$.\footnote{In fact in \citet{barlow-85} the inequality is written
  $\ll$ rather than $\gg$, but it is clear the latter  was intended.
  The same paper was also published as \citet{barlow-87}; there the
  inequality is the correct way round.}
  Any detective knows that, for a coincidence to be suspicious, the
  events themselves must be rare ones, and that if they are rare
  enough, even a   single occurrence is significant.
\end{quote}

\citet{edelman-hiles-yang-intrator-02} refer to the \emph{principle of
suspicious coincidences} as where ``two candidate fragments $A$ and
$B$ should be combined into a composite object if the probability of
their joint appearance $P(A,B)$ is much higher than $P(A) P(B)$ ...''

\sgap

The fundamental problem here is to detect if there is a significant
\emph{association} between events $A$ and $B$. This can arise in many
different contexts, such as:
\begin{itemize}
\item an animal detecting that eating a certain plant is associated
  with subsequent illness;
\item detecting that a certain drug is associated with a particular
  adverse drug reaction;
\item detecting the association between a visual stimulus that contains
  an image of the subject's grandmother or not, and the response of a
  putative ``grandmother cell'';
\item detecting that particular successive words in text are
  associated more frequently than by chance---this is called a
  ``collocation'', an example being the bigram ``carbon dioxide'';
\item a geneticist determining that two genes are in linkage
  disequilibrium (see e.g., \citealt*{lewontin-64});
\item detecting that the pattern of two edges in a visual scene making
  a corner junction occur more frequently than by chance.
\end{itemize}  

Below we review various \emph{measures of association} from the
literature, notably Yule's $Y$ \citep{yule-1912}, which depends solely
on the odds ratio and is invariant to the marginal distributions of
the two variables. We then discuss measures of association based on
the mutual information and pointwise mutual information, which make
use of the ratio $P(A,B)/P(A)P(B)$, as proposed by Barlow and others
across diverse literatures.  Finally we consider the pros and cons of
using PMI to flag up suspicious coincidences, and discuss its estimation from
data when (some of) the counts in the table are low.

\section*{$2 \times 2$ Contingency Tables}
Consider two random variables $X$ and $Y$ that take on values of 0 or
1. The $2 \times 2$ contingency table has the form
\begin{equation}
  P = \left(\!\begin{array}{cc}
             p_{00} & p_{01} \\
             p_{10} & p_{11}
  \end{array}\!\right),
  \label{eq:Ptable}
\end{equation}
where, for example, $p_{01} = p(X=0, Y=1)$. We will also say the event
$x$ occurs if $X=1$, and similarly for $y$. We denote the marginals
with ``dot'' notation, so that e.g.\ $p_{1 \cdot} = p(X=1) = p_{10} + p_{11}$.

$P$ is defined by 3 degrees of freedom (as the entries sum to 1). Two
of these are taken up by the marginals, leaving one additional degree
of freedom. Given a table $P$ we can manipulate the marginals by
multiplying the rows and columns with positive numbers and
renormalizing. Such a transformation is shown, e.g.,
in \citet[eq.\ 1]{hasenclever-scholz-16}, viz,\
\begin{equation}
g_{\mu,\nu}(P) = \frac{1}{Z(\mu,\nu)}  \left(\!\begin{array}{cc}
             \mu \nu p_{00} & \mu p_{01} \\
             \nu p_{10} & p_{11}
           \end{array}\!\right) ,
\end{equation}
where $Z(\mu,\nu) = \mu \nu p_{00} + \mu p_{01} + \nu p_{10} +
p_{11}$. The odds ratio
\begin{equation}
\lambda = \frac{p_{00} p_{11}}{p_{01} p_{10}}
\end{equation}
can be seen to be invariant to the action of this
margin manipulation transformation, and thus defines the third degree
of freedom. An odds ratio of 1 implies that there is no association, and that $P$
is equal to the product of the marginals.

The ``canonical'' table with marginals of $1/2$ but with the same odds ratio as $P$
is given by 
\begin{equation}
  P_{can} = \left(\!\begin{array}{cc}
             \frac{\rlam}{2(1 + \rlam)} &   \frac{1}{2(1 + \rlam)} \\
             \frac{1}{2(1 + \rlam)} & \frac{\rlam}{2(1 + \rlam)} 
           \end{array}\!\right), \label{eq:Pcan}
\end{equation}
as shown by \citet{yule-1912}.
Like a copula for continuous variables, this allows a separation of the
marginals from the dependence structure between $X$ and $Y$.

The table $P$ can also be expressed in terms of a deviation from the
product of the marginals (see e.g.\
\citealt*[p.\ 24]{hasenclever-scholz-16}) as 
\begin{equation}
  P = \left(\!\begin{array}{cc}
             p_{0 \cdot} p_{\cdot 0} + D & p_{0 \cdot} p_{\cdot 1} - D \\
             p_{1 \cdot} p_{\cdot 0} - D & p_{1 \cdot} p_{\cdot 1} + D 
           \end{array}\!\right),  \label{eq:marg+D}
\end{equation}
where $D = p_{00} p_{11} - p_{01} p_{10} = p_{11} - p_{1 \cdot}
p_{\cdot 1}$ etc. In genetics $D$ is known as the coefficient of
linkage disequilibrium for two genes.

\paragraph{Estimation from Data: }
Eq.\ \ref{eq:Ptable} is given in terms of probabilities such as
$p_{01}$. However, observational data does not directly provide such
probabilities, but counts associated with the corresponding cells. The
maximum likelihood estimator (MLE) for $p_{ij}$ is, of course,
$n_{ij}/n$, where $n_{ij}$ is the count associated with cell $ij$, and
$n$ is the total number of counts. The MLE has well-known issues when
(some of) the counts are small. Bayesian approaches to address this
are discussed below in the section headed Detecting Associations with
Pointwise Mutual Information.

\section*{Classical Measures of Association}
For two Gaussian continuous random variables, there is a natural
measure of their association, the correlation coefficient. This is
independent of the individual (marginal) variances of each variable,
and lies in the interval $[-1,1]$.

For the $2 \times 2$ table many measures of association have been
devised. One such is Yule's $Y$ \citep{yule-1912}, where
\begin{equation}
  Y = \frac{\rlam -1}{\rlam + 1}.
\end{equation}
Like the correlation coefficient, $Y$ also lies in the range of
$[-1,1]$, with a value of 0 reflecting that there is no
association. Its dependence only on $\lambda$ means that it is
invariant to the marginals in the table. $Y(1/\lambda) = -
Y(\lambda)$, so $Y$ is an odd function of $\log(\lambda)$.
\citet{edwards-63} argued that measures of association must be
functions of the odds ratio.

There are a number of desirable properties for a  \emph{measure of
association} $\eta$ between binary variables.
For example \citet[p.\ 22]{hasenclever-scholz-16} list:
\begin{itemize}
\item[(a)]  $\eta$ is zero on independent tables.
\item[(b)]  $\eta$ is a strictly increasing function of the odds-ratio
  when restricted to tables with fixed margins.
\item[(c)] $\eta$ respects the symmetry group $D_4$, namely (1) $\eta$
  is
  symmetric in the markers, i.e.\ invariant to matrix transposition,
  and (2)
  $\eta$ changes sign when the states of a marker are transposed (row
  or column   transposition).
\item[(d)] The range of the function is restricted to $(-1  ,1)$.
\end{itemize}  

As well as Yule's $Y$,\footnote{Yule had earlier proposed $Q =
(\lambda -1)/ (\lambda +1)$ as a measure of association, but his
discussion on p.\ 592 of \citet{yule-1912} gives a number of reasons
for preferring $Y$ to $Q$.} several other measures of association have
been proposed; indeed \citet{tan-kumar-srivastava-04} list 21. 
Other measures of association include Lewontin's $D'$
\cite{lewontin-64}, which standardizes $D$ from eq.\ \ref{eq:marg+D}
by dividing it by the maximum value it can take on, which depends on
the marginals of the table; and the binary correlation coefficient $r$
which standardizes $D$ by $\sqrt{p_{0 \cdot}p_{\cdot 0} p_{1 \cdot}
  p_{\cdot 1}}$. For the canonical table, it turns out that $D' = r = Y$.

\section*{Information Theoretic Measures of Association}
Barlow's definition of a suspicious coincidence suggests consideration
of the quantity
\begin{equation}
  i(x,y) = \log \frac{p(x,y)}{p(x)p(y)} .
\end{equation}
Indeed $i(x,y)$ has been proposed in different literatures; for example
\citet{church-hanks-90} studied it for word associations in
linguistics. $i(x,y)$ is termed the 
\emph{pointwise mutual information} (PMI), e.g.\ in the statistical
natural language processing textbook of \citet{manning-schuetze-99}.
In pharmacovigilance, \citet{bate-etal-98} call $i(x,y)$ 
the \emph{information component} (IC), as it is one component of the
mutual information calculation in a $2 \times 2$ table, and it is also
studied in \citet{dumouchel-99}. And in the data mining literature
\citet{silverstein-brin-motwani-98} define the \emph{interest} to
be the ratio $p(x,y)/(p(x)p(y))$ (i.e.\ without the $\log$). 

Note that while $Y$, $D'$ and $r$ consider the \emph{difference} $D =
p_{11} - p_{1 \cdot} p_{\cdot 1} = p(x,y) - p(x) p(y)$, $i(x,y)$
considers the log \emph{ratio} of these terms. Thus $i(x,y)$ considers
the ratio of the observed and expected probabilities for the event
$(x,y)$, where the expected model is that of independence.

The mutual information (MI) is defined as 
\begin{equation}
I(X;Y) = \sum_{i,j \in \{0, 1\} } p(X=i,Y=j) \log \frac{P(X=i,Y=j)}{P(X=i) P(Y=j)} .
\label{eq:midefn}
\end{equation}
We have that $I(X;Y) \ge 0$, with $I(X;Y) = 0$ when $X$ and $Y$ are
independent.

Both PMI and MI as defined above depend on the marginal probabilities
in the table. To see this, use $p(x,y)\le p(x)$ or $p(x,y)\le p(y)$,
so $i(x,y) \le \min( - \log p(x), \; - \log p(y))$, i.e.\ favouring
``sparsity'' (low probability).  The MI is maximal for a diagonal (or
anti-diagonal) table with marginals of $1/2$, the opposite trend to
PMI.

There have been various proposals to normalize the PMI and MI to make
them fit in the range $[-1,1]$ and $[0, 1]$ respectively. For example
\citet{bouma-09} defined the normalized PMI (NPMI) as $i_n(x,y) =
i(x,y)/h(x,y)$ for $p(x,y) > 0$, where $h(x,y) = - \log p(x,y)$. NPMI
ranges from $+1$ when events $x$ and $y$ only occur together, through
0 when they are independent, to $-1$ when $x$ and $y$ occur separately
but not together.  Similarly there are a number of proposals for
normalizing the mutual information; \citet{bouma-09} suggests
$I_n(X;Y) = I(X;Y)/H(X,Y)$, where $H(X,Y)$ is the joint entropy of $X$
and $Y$. $I_n(X;Y)$ (termed the normalized MI or NMI) takes on a value
of $+1$ if $X$ and $Y$ are perfectly associated, and 0 if they are
independent. Alternative normalizations of the MI by $H(X)$ or $H(Y)$
have also been proposed, these are termed uncertainty coefficients in
\citet[sec.\ 14.7.4]{press-teukolsky-vetterling-flannery-07}.  NMI is
not strictly a measure of association as defined above, as it does not
take on negative values, but following the construction in
\citet[p.\ 24]{hasenclever-scholz-16}, one can e.g.\ define the
\emph{signed} NMI as $\mathrm{sign}(D) I_n(X;Y)$.

Given that the canonical table removes the effect of the marginals, it
is natural to consider PMI and MI as a function of $\lambda$.
Using the canonical table from eq.\ \ref{eq:Pcan}, we obtain
\begin{equation}
  i_{\lambda}(x,y) = \log \frac{2 \rlam}{1 + \rlam},
  \label{eq:ilambda}
\end{equation}
which takes on a value of 0 for $\lambda = 1$ (independence), and tends
to a value of $\log(2)$ as $\lambda$ tends to infinity. For
$\lambda < 1 $ the value of $i_{\lambda}(x,y)$ becomes negative and
diverges to $- \infty$ as $\lambda \rightarrow 0$. However, by
studying the canonical table it would make more sense in this case to consider one
of the ``anti-diagonal'' cells in $P_{can}$ which will have a
probability greater than $1/2$ as the ``event''. In general we can
treat all four cells of the contingency table as the ``joint event'', 
compute the PMI for each, and return the maximum. For the canonical table
with $\lambda < 1 $ this means that we transform
$\lambda \rightarrow 1/\lambda$ and compute $i_{\lambda}$ as per eq.\ \ref{eq:ilambda}.

For the MI of the canonical table, we obtain (after some manipulation)
\begin{equation}
I_{\lambda}(X;Y) =  \frac{\rlam}{1 + \rlam} \log \rlam - \log (1 +
\rlam) + \log 2 .
\end{equation}
Analysis of $I_{\lambda}(X;Y)$ shows that it is invariant if we
transform $\lambda$ to $1/\lambda$, so a plot of $I_{\lambda}(X;Y)$
against $\log(\lambda)$ is symmetric around 0, and tends to the value
$\log(2)$ as $\lambda$ tends to 0 or infinity.

\begin{figure*}
\begin{center}
\includegraphics[width=4in]{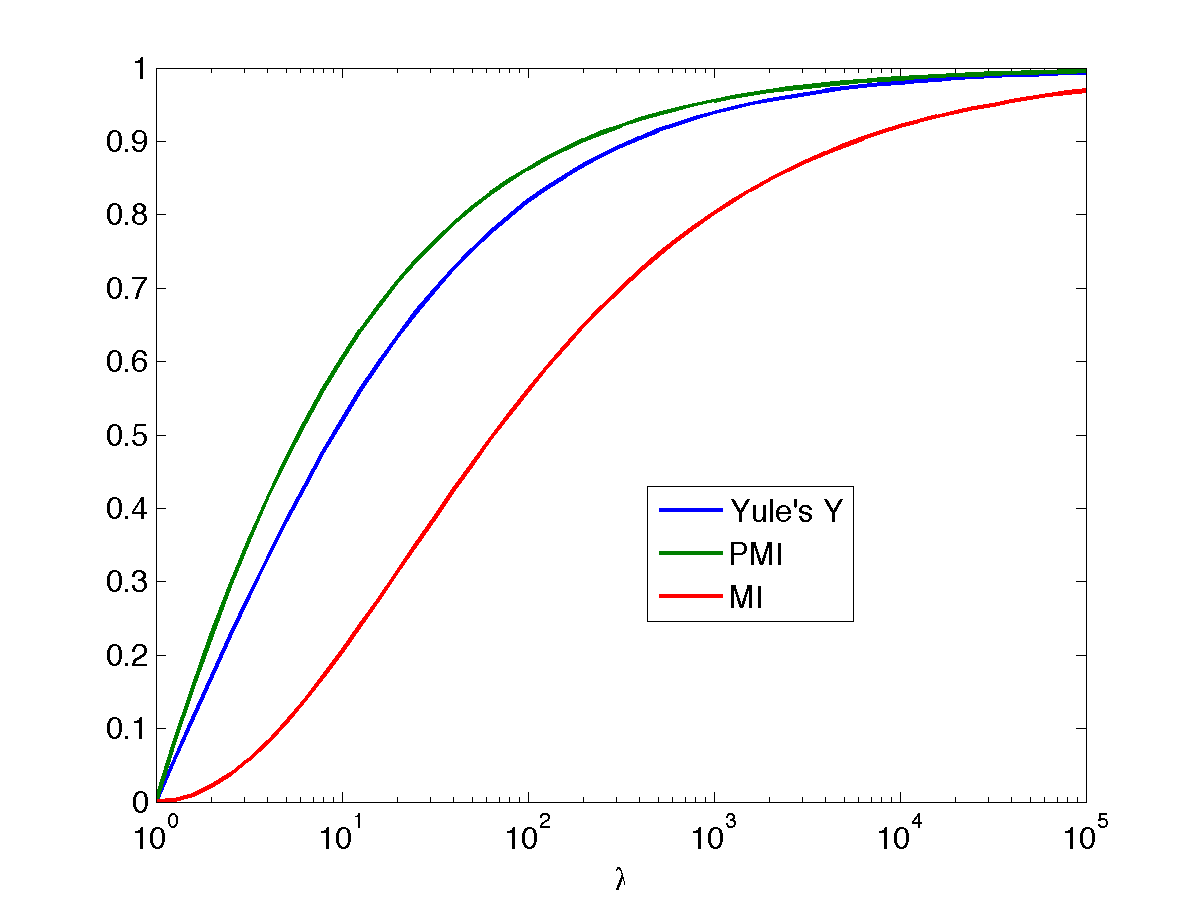}
\end{center}  
\caption{Plots of $Y$, $i_{\lambda}$ (PMI) and  $I_{\lambda}$ (MI) against
  $\lambda$ (log scale) for $\lambda \ge 1$. \label{fig:YiI}}
\end{figure*}

Plots of $Y$, $i_{\lambda}$ and $I_{\lambda}$ for $\lambda \ge 1$ in
Figure \ref{fig:YiI} show a
similar behaviour, monotonically increasing to a maximum value as
$\lambda \rightarrow \infty$. If we choose logs to base 2, then the
maximum value is 1 in all three cases. As $Y$ is already
well-established (since 1912!), it does not seem necessary to
promote $i_{\lambda}$ or $I_{\lambda}$ as alternatives, when
considering the canonical table.

\begin{table}
{\small
\begin{center}    
\begin{tabular}{ccc}

\begin{tabular}{lcc|c}
             & recover & die & marginals \\
vaccinated   & 0.840  & 0.043  & 0.883 \\
unvaccinated & 0.059  & 0.058  & 0.117 \\ \hline
marginals    &  0.899  & 0.101   
  \end{tabular}

&
  
 \begin{tabular}{cc|c}
      &   & \\
 0.476  & 0.024  & 0.500 \\
 0.252  & 0.248  & 0.500 \\ \hline
 0.728  & 0.272  &    
 \end{tabular}

   &

 \begin{tabular}{cc|c}
      &   & \\
 0.408  & 0.092  & 0.500 \\
 0.092  & 0.408  & 0.500 \\ \hline
 0.500  & 0.500  & 
 \end{tabular}

 \\
PMI = 2.300, MI = 0.108 &
PMI = 0.866, MI = 0.205 & 
PMI = 0.705, MI = 0.310 \\

(a) original table & 
(b) vaccination rate 50\% &
(c) canonical table 
\end{tabular}  
\end{center}
} 

\caption{$2 \times 2$ contingency tables for the association
between vaccination and death from smallpox. (a) is the original
table based on the data in \citet{yule-1912}, (b) adjusts the marginals
for vaccinated/unvaccinated to be 50/50, and (c) is the canonical
table where the marginals are both 50/50. In all three tables, Yule's
$Y = 0.630$. \label{tab:vaccunvacc}
}
\end{table}  

\section*{Detecting Associations with Pointwise Mutual Information}
As we have seen, the raw PMI score is not invariant to the
distribution of the marginals. This can be seen in Table
\ref{tab:vaccunvacc}, which concerns the association between
vaccination and death from smallpox; the original proportions in panel
(a) are based on the Sheffield data in Table I of
\citet{yule-1912}. In panel (b) the marginals of the table wrt
vaccination have been adjusted to 50/50 (as may have happened if these
data had been collected in a randomized controlled trial), and in
panel (c) we have the canonical table where both marginals are
50/50.\footnote{Yule (1912) comments that on the canonical table that
``These are, of course, not the actual proportions, but the
proportions that would have resulted if an omnipotent demon of
unpleasant character (no relation of Maxwell's friend) could have
visited Sheffield [...], and raised the fatality rate and the
proportion of unvaccinated [...] to 50 per cent without otherwise
altering the facts.''}  Notice that the PMI is highest for the
original (unbalanced) table, and decreases as the marginals are
balanced. Conversely the MI is lowest in the the original (unbalanced)
table, and increases as the marginals are balanced. Of course Yule's
$Y$ is constant throughout, by construction.

As another example, consider fixing $\lambda$ but adjusting the marginal
probabilities of events $x$ and $y$. For example, for
$\lambda=16$, PMI takes on the values of 0.678, 1.642, 2.293, 3.642
and 3.958 (using
logs to base 2) as $p(x)=p(y)$ varies from 0.5, 0.2, 0.1, 0.01 and 0.001.
This is particularly problematic as low counts will give rise to
uncertainty in the estimation of the required probabilities
(especially of the joint event). In the context of word associations,
\citet[sec.\ 5.4]{manning-schuetze-99} argue that PMI ``does not
capture the intuitive notion of an interesting collocation very
well'', and mention work which multiplies it by $p(x,y)$ as one
strategy to compensate for the bias in favour of rare events.

On the other hand, \citet{barlow-85} suggested that sparsity is
important for the detection of suspicious coincidences, i.e.\ that
``the events themselves must be rare ones''. It is true that a low
$p(y)$ gives more ``headroom'' for the ratio $p(y|x)/p(y)$ to be
large. The PMI score is used extensively in pharmacovigilance, where
the aim is to detect associations between drugs taken and adverse drug
reactions (ADRs).  In this context, the ratio $p(x,y)/p(x)p(y) =
p(y|x)/p(y)$ is termed the \emph{relative reporting ratio} (RRR), and
compares the relative probability of an adverse drug reaction $y$
given treatment with drug $x$, compared to the base rate
$p(y)$. Another commonly used measure is the \emph{proportional
reporting ratio} (PRR), defined as $p(y|x)/p(y|\neg x)$.
The US Food and Drug Administration (FDA) white paper
\citep{duggirala-etal-18} describes the use of both RRR and PRR for
detecting ADRs in routine surveillance activities.

Above we have described maximum likelihood estimation for the
probabilities in the $2 \times 2$ table, based on counts. However,
there are well-known issues with the MLE when (some of) the counts are
small. This naturally suggests a Bayesian approach, and there is a 
considerable literature
on the Bayesian analysis of contingency tables, as reviewed e.g.\ in
\citet{agresti-13}. There are different sampling models
depending on how the data is assumed to be generated, as described in
\citet[sec. 2.1.5]{agresti-13}. If all 4 counts
are unrestricted, a natural assumption is that each $n_{ij}$ is drawn from
a Poisson distribution with mean $\mu_{ij}$, which can be given a
Gamma prior. Alternatively if $n$ is
fixed, the sampling model is a multinomial, and the conjugate prior
is a Dirichlet distribution. If one set of marginals is fixed, then
the data is drawn from two Binomial distributions, each of which can be
given a Beta prior. If both marginal totals are fixed, this
corresponds to Fisher's famous ``Lady Tasting Tea'' experiment, and
the sampling distribution of any cell in the table follows a
hypergeometric distribution.  Section 3.6 of \citet{agresti-13} covers
Bayesian inference for two-way contingency tables, and
\citet{agresti-min-05} discuss Bayesian confidence intervals for
association parameters, such as the odds ratio.

\citet{dumouchel-99} applied an Empirical Bayes approach to
consider sampling variability for PMI (aka RRR) 
in the context of adverse drug reactions. He assumed that
each $n_{11}$ is a draw from a Poisson distribution with unknown
mean $\mu_{11}$, and that the object of interest is $\rho_{11} =
\mu_{11}/E_{11}$, where $E_{11}$ is the expected count (assumed known)
under the assumption that the variables
are independent. Using a mixture of Gamma distributions prior for
$\rho_{11}$, DuMouchel obtained the posterior mean
$E[\log(\rho_{11})|n_{11}]$,
rather than just considering the sample estimate $n_{11}/E_{11}$.
The mixture prior was used to express the belief that when testing 
many associations,  most will have a PMI of near 0, but there
will be some that have significantly larger values. This method is
known as the Multi-Item Gamma Poisson Shrinker (MGPS). The value of
this approach is that
Bayesian shrinkage corrects for the high variability in the RRR
sample estimate $n_{11}/E_{11}$  that results from small counts.

\section*{Summary}
Motivated by Barlow's hypothesis about suspicious coincidences, we
have reviewed the properties of $2 \times 2$ contingency tables for
association analysis, with a focus on the odds ratio $\lambda$ and
Yule's $Y$.  We have considered the mutual information and pointwise
mutual information as measures of association, along with normalized
versions thereof. We have shown that, considered as functions of
$\lambda$ in the canonical table, MI and PMI behave similarly to $Y$
for $\lambda \ge 1$, increasing monotonically with $\lambda$ (and can
be made similar for $0 < \lambda < 1$).

As well as $Y$, the PMI measure $i(x,y) = \log p(x,y)/(p(x)p(y)$ can
also be used to identify suspicious coincidences, and it is used in
practice, for example, in pharmacovigilance.  We have discussed the
pros and cons of using it in this way, bearing in mind the sensitivity
of the PMI to the marginals, with increased scores for sparser events.
When some of the counts in the table are low, Bayesian approaches can
be useful for the estimation of PMI from raw counts.

\section*{Acknowledgments}
I thank Peter Dayan and Iain Murray for helpful comments on an early
draft of the paper, and the anonymous reviewers whose comments
helped to improve the paper.

\bibliographystyle{apalike}

\section*{Addendum: Earlier Work on Pointwise Mutual Information}
Subsequent to the publication of this paper, I became aware of the
paper by \citet{good-56}, which terms pointwise mutual information the
``association factor''. In this paper on p.\ 114 Good mentions a
number of earlier references to this quantity. I have been able to
trace discussion of PMI to \citet[p.\ 53]{woodward-53} (where it is
termed the ``information transfer between x and y''), and (without the
log) to \citet[p.\ 170]{keynes-21} (where it is termed the ``coefficient of
dependence''). For the latter reference, the paper by
\citet{brossel-15} was helpful in tracking down the citation.

\end{document}